\documentclass[letterpaper, 10 pt, conference]{ieeeconf}  %

\IEEEoverridecommandlockouts                              %
        
\makeatletter
\let\labelindent\relax
\makeatother
\usepackage{enumitem}
\usepackage{cite}
\usepackage{amsmath,amssymb,amsfonts}

\usepackage{arydshln}
\usepackage{graphicx}
\usepackage{textcomp}
\usepackage{xcolor}
\usepackage{arydshln}
\usepackage{multirow}
\usepackage{placeins}
\def\BibTeX{{\rm B\kern-.05em{\sc i\kern-.025em b}\kern-.08em
    T\kern-.1667em\lower.7ex\hbox{E}\kern-.125emX}}
\usepackage{mathtools}
\usepackage{algorithm}
\usepackage{algcompatible}
\usepackage{algpseudocode}
\usepackage{subcaption}
\usepackage{soul}
\usepackage{afterpage}

\usepackage{siunitx}
\usepackage{url}
\usepackage{overpic}
\usepackage{pgf}
\usepackage{microtype}
\usepackage{booktabs}
\usepackage{color, colortbl}  %
\usepackage{hyperref}

\usepackage{subcaption}
\newcolumntype{C}[1]{>{\centering\let\newline\\\arraybackslash\hspace{0pt}}m{#1}}
\newcolumntype{R}[1]{>{\raggedleft\let\newline\\\arraybackslash\hspace{0pt}}m{#1}}

\title{\LARGE \bf PlanarMesh: Building Compact 3D Meshes from LiDAR using Incremental Adaptive Resolution Reconstruction}

\author{
    Jiahao Wang$^{1}$, Nived Chebrolu$^{1}$, Yifu Tao$^{1}$, Lintong Zhang $^{2}$, Ayoung Kim$^{3}$, and Maurice Fallon$^{1}$
    \thanks{
        $^1$Oxford Robotics Institute, Univ. of Oxford, UK. 
        \{
        \href{mailto:jiahaowang@robots.ox.ac.uk}{\nolinkurl{jiahaowang}}, \href{mailto:yifu@robots.ox.ac.uk}{\nolinkurl{yifu}}, \href{mailto:nived@robots.ox.ac.uk}{\nolinkurl{nived}}, \href{mailto:mfallon@robots.ox.ac.uk}{\nolinkurl{mfallon}}
        \} \texttt{@robots.ox.ac.uk}
        \newline
        \indent $^2$NavLive Ltd, Oxford, UK. \href{mailto:lzhang@navlive.ai}
        {\nolinkurl{lzhang@navlive.ai}}
        \newline
        \indent $^3$Seoul National University, South Korea. \href{mailto:ayoungk@snu.ac.kr}{\nolinkurl{ayoungk@snu.ac.kr}}
        \newline
        This project has been partly funded by the National Research Foundation of Korea (NRF) grant funded by the Korea government (MSIT)(No. RS-2024-00461409), NavLive Ltd (Wang), and Royal Society Univ. Research Fellowship (Fallon). For the purpose of open access, the authors have applied a Creative Commons Attribution (CC BY) license to any Accepted Manuscript version arising.
        }
}

\begin{document}

\newcommand{\compressionRatio}{10\ }
\newcommand{\processingSpeed}{2\ }

\newcommand{\textPlanarMesh}{\text{planar-mesh}}
\newcommand{\textAlgorithmName}{\text{PlanarMesh}}
\newcommand{\textLidar}{\text{LiDAR}\ }

\newcommand{\lidarPoint}{\mathbf{l}_{p}}
\newcommand{\lidarOrigin}{\mathbf{o}}
\newcommand{\lidarRay}{\hat{\mathbf{l}}}

\newcommand{\radius}{r}
\newcommand{\radiusBoundary}{\radius_{\partial}}

\newcommand{\planePosition}{\mathbf{p}}
\newcommand{\planeNormal}{\hat{\mathbf{n}}}

\newcommand{\plane}{P}
\newcommand{\planes}{\mathbf{\plane}}
\newcommand{\mesh}{M}
\newcommand{\meshes}{\mathbf{mesh}}
\newcommand{\planarmesh}{\mesh_{\plane}}
\newcommand{\planarmeshes}{\mathbf{\planarmesh}}
\newcommand{\map}{\mathcal{\planarmesh}}
\newcommand{\planarmeshFront}{\planarmesh^{\text{front}}}
\newcommand{\planarmeshWithin}{\planarmesh^{\text{within}}}
\newcommand{\planarmeshBehind}{\planarmesh^{\text{behind}}}
\newcommand{\planarmeshSeed}{\planarmesh^{\text{seed}}}

\newcommand{\planePoints}{\mathcal{L}_{\plane}}

\newcommand{\range}{d}
\newcommand{\expectedDistance}{\mu_{\range}}
\newcommand{\stdDistance}{\sigma_{\range}}

\newcommand{\faces}{F}
\newcommand{\face}{f}
\newcommand{\edges}{E}
\newcommand{\edge}{e}
\newcommand{\vertices}{V}
\newcommand{\vertex}{v}
\newcommand{\boundaryVertices}{V_{\partial}}

\newcommand{\FISTree}{\mathcal{T}_{\faces}}
\newcommand{\RRSTree}{\mathcal{T}_{\boundaryVertices}}

\newcommand{\FISFunc}{\texttt{FIS}}
\newcommand{\RRSFunc}{\texttt{RRS}}
\newcommand{\planarMeshOf}{\texttt{PM}}

\newcommand{\intersectingFaces}{\faces^{*}}
\newcommand{\includingVertices}{\boundaryVertices^{*}}

\newcommand{\planarMeshesOfFaces}{\planarmeshes_{\intersectingFaces}}
\newcommand{\planarMeshesOfVertices}{\planarmeshes_{\includingVertices}}

\newcommand{\planarMeshesOfFacesWithin}{\planarMeshesOfFaces^{\text{within}}}
\newcommand{\planarMeshesOfFacesBehind}{\planarMeshesOfFaces^{\text{behind}}}
\newcommand{\planarMeshesOfVerticesWithin}{\planarMeshesOfVertices^{\text{within}}}
\newcommand{\planarMeshesOfVerticesSeed}{\planarMeshesOfVertices^{\text{seed}}}

\newcommand{\updatedPlanarMesh}{\tilde{\mesh}_{\plane}}
\newcommand{\updatedPlanarMeshes}{\tilde{\mathbf{\mesh}}_{\mathbf{\plane}}}
\newcommand{\updatedMap}{\tilde{\mathcal{\mesh}}_{\mathcal{\plane}}}
\newcommand{\updatedPlanarMeshesOfFaces}{{\updatedPlanarMeshes}{}_{\intersectingFaces}}
\newcommand{\updatedPlanarMeshesOfVertices}{{\updatedPlanarMeshes}{}_{\includingVertices}}
\newcommand{\updatedPlanarMeshesOfFacesBehind}{\updatedPlanarMeshesOfFaces^{\text{behind}}}
\newcommand{\newSeedPlanarMesh}{\updatedPlanarMesh^{\text{seed}}}

\def\secref#1{Sec.~\ref{#1}}
\def\figref#1{Fig.~\ref{#1}}
\def\tabref#1{Tab.~\ref{#1}}
\def\eqref#1{Eq.~(\ref{#1})}
\def\algref#1{Alg.~\ref{#1}}

\makeatletter
    \let\@oldmaketitle\@maketitle%
    \renewcommand{\@maketitle}
    {
        \@oldmaketitle%
        \bigskip
        \centering
        \includegraphics[width=\textwidth]{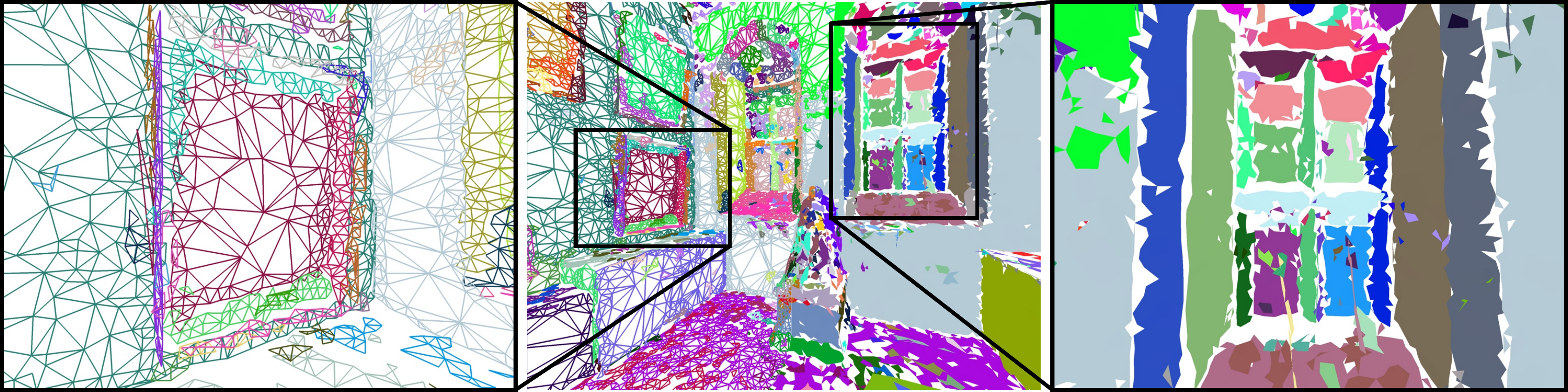}
        \captionof{figure}{PlanarMesh can reconstruct buildings and indoor spaces as a set of detailed, yet minimal, planar surfaces in real-time (\href{https://www.youtube.com/watch?v=GmMR96nYp90}{link to video}). It uses local curvature estimate to achieve adaptive resolution surface meshing. Left to right: [l] Plane reconstruction of a painting frame, [c] of a wider room, [r] and of inset window frame.}
        \label{fig:header}   
    }
\makeatother
\maketitle
\setcounter{figure}{1}

\begin{abstract}
Building an online 3D LiDAR mapping system that produces a detailed surface reconstruction while remaining computationally efficient is a challenging task. In this paper, we present PlanarMesh, a novel incremental, mesh-based LiDAR reconstruction system that adaptively adjusts mesh resolution to achieve compact, detailed reconstructions in real-time. It introduces a new representation, planar-mesh, which combines plane modeling and meshing to capture both large surfaces and detailed geometry. The planar-mesh can be incrementally updated considering both local surface curvature and free-space information from sensor measurements. We employ a multi-threaded architecture with a Bounding Volume Hierarchy (BVH) for efficient data storage and fast search operations, enabling real-time performance. Experimental results show that our method achieves reconstruction accuracy on par with, or exceeding, state-of-the-art techniques—including truncated signed distance functions, occupancy mapping, and voxel-based meshing—while producing smaller output file sizes (10 times smaller than raw input and more than 5 times smaller than mesh-based methods) and maintaining real-time performance (around 2 Hz for a 64-beam sensor).
\end{abstract}

\section{Introduction}

Mobile 3D LiDAR scanners are becoming more affordable and reliable. They are now widely used in a variety of robotics and geospatial applications to model environments and buildings and provide accurate and reliable spatial data. However, the immense volume of raw point cloud data generated by these sensors poses a significant challenge when processing, storing, and carrying out real-time 3D reconstruction. An efficient reconstruction system is essential for converting raw LiDAR measurements into a compact and detailed representation that can be used for downstream tasks such as mission planning, localization, and inspection. In addition, when mapping online, sensor measurements arrive in a continuous stream. This requires the system to operate incrementally and maintain an up-to-date 3D model without imposing excessive computational overhead. 

Triangle mesh is a popular 3D surface representation for its compactness and explicit surface connectivity, and there exist systems such as ImMesh~\cite{lin_immesh_2023}, SLAMesh~\cite{ruan_slamesh_2023}, and IDTMM~\cite{niedzwiedzki_idtmm_2023} that can perform incremental mesh reconstruction. However, these methods create meshes with fixed resolution, which limits their performance: at low resolution, meshes are over-smoothed and cannot capture fine geometric details, whereas at high resolution, the increased number of mesh elements leads to larger file sizes, and smaller mesh elements that are more susceptible to over-fitting noise.

We introduce PlanarMesh, a novel surface reconstruction system that incrementally generates compact and detailed 3D reconstruction. PlanarMesh leverages the observation that planar structures (e.g. walls, ceilings, ground) are ubiquitous in built environment, and achieves adaptive resolution reconstruction by combining plane modeling with meshing. It uses planar models to represent flat surfaces, which provide an averaging effect to individual measurement noise, while employing variable-resolution meshing to capture fine details where needed. This adaptability is guided by local curvature information, which determines the appropriate resolution during mesh expansion. To efficiently update the planar-mesh, we propose Reverse Radius Search (RRS) for curvature-informed neighbor search and Face Intersection Search (FIS) to incorporate free-space information from LiDAR measurements, improving reconstruction accuracy. Our experimental results demonstrate that PlanarMesh achieves the highest compression rates, reducing mesh sizes by more than 5 times compared to existing mesh-based methods, while also maintaining comparable reconstruction quality.

In summary, our key contributions are as follows:
\begin{itemize}
    \item \textbf{Adaptive Mesh Resolution:} We introduce PlanarMesh, which enables adaptive reconstruction by combining plane modeling with meshing. 
    \item \textbf{Reduced Map Size Footprint:} Our adaptive meshing significantly reduces file sizes, achieving \compressionRatio times compression over raw point clouds and more than 5 times smaller outputs than existing mesh-based methods. 
    \item \textbf{Incremental Mesh Reconstruction}: PlanarMesh updates existing planar-meshes as well as search trees incrementally, leveraging both local curvature and free-space information for an accurate reconstruction. 
    \item \textbf{Real-Time Performance:} We achieve near LiDAR frame-rate reconstruction through a custom Bounding Volume Hierarchy (BVH) implementation, supporting efficient incremental updates and multi-threaded queries.
\end{itemize}

\section{Related Works}
\label{sec:related_works}

We categorize the related work according to the 3D reconstruction approach taken so as to highlight the strengths and limitations of each, and to show how our work builds upon or differs from existing methods. These categories include point-based, surfel-based, mesh-based, occupancy-based, implicit surfaces, and global interpolation methods.

\textbf{Point-based:}
Early work in 3D reconstruction began with point-based representations \cite{zhang_loam_2014}, where raw point cloud data formed the basis for subsequent processing. However, point clouds by themselves lack structural connectivity, which limits their utility in many applications.

\textbf{Surfel-based:}
Surfels enhance point-based representations by incorporating local surface orientation and size properties, allowing for more descriptive scene reconstruction \cite{pfister_surfels_2000}. Systems like ElasticFusion \cite{whelan_elasticfusion_2015} and Probabilistic Surfel Fusion \cite{park_probabilistic_2017} benefit from this discrete surface representation—ElasticFusion enables elastic deformation upon loop closure for greater flexibility, while Probabilistic Surfel Fusion employs surfel merging to improve robustness to noise. However, surfel-based methods suffer from a lack of explicit connectivity between surfels. 
Our system strikes a middle ground by using individual planar-mesh elements that can be integrated into future loop closure systems, fitting planes to points to reduce noise, and leveraging mesh structures to address the connectivity issues seen in surfel methods.

\textbf{Meshes:}
In general, mesh-based methods can be divided into two subcategories: volumetric and surface-based \cite{boissonnat_geometric_1984}. Volumetric methods, such as Alpha Shapes \cite{Bernardini1997SamplingAR} and the Crust Algorithm \cite{amenta_new_1998, amenta_power_2001}, compute exhaustive triangulations—often enforcing Delaunay properties—and then selectively remove elements to extract the correct mesh geometry. Although effective for post-processing, these methods struggle with incremental updates, as repeated global triangulations are computationally intensive. 

In contrast, surface-based methods (often related to region growing techniques \cite{bernardini_ball-pivoting_1999, mencl_graphbased_1995}) rely on local tangent plane approximations to sequentially build mesh surfaces from a set of points. Although they can locally expand, these approaches were originally designed for offline processing and lack robust mechanisms to seamlessly integrate new points. 
ImMesh \cite{lin_immesh_2023} and SLAMesh \cite{ruan_slamesh_2023} address these problems by creating mesh surfaces per voxel, and incrementally update the mesh inside each voxel. However, both approaches impose a maximum mesh size, leading to potential discontinuities between adjacent voxels. IDTMM \cite{niedzwiedzki_idtmm_2023}, on the other hand,  achieves incremental updates using Kalman filtering on mesh vertex positions not limited by voxel size. 
While our method shares the ability to operate without voxel size constraints like IDTMM, we further differentiate ourselves by supporting adaptive resolution, allowing efficient representation of both large planar regions and fine details. 

\textbf{Occupancy Maps:}
Occupancy mapping methods such as OctoMap \cite{hornung_octomap_2013}, SuperEight \cite{vespa_efficient_2018}, and D-Map \cite{cai_occupancy_2024} store occupancy probabilities. These methods inherently leverage free-space information from LiDAR, offering robustness in dynamic environments. SuperEight and Wavemap \cite{reijgwart2023wavemap} use multi-resolution grids to achieve adaptive resolution, which helps to represent complex scenes with a smaller data footprint. However, occupancy mapping methods lack explicit surface representation, requiring post-processing techniques such as Marching Cubes \cite{lorensen_marching_1987} to extract surfaces, which can introduce geometric smoothing artifacts. Additionally, they struggle to accurately represent large planar surfaces when the plane is not aligned with the grid direction, often resulting in staircase-like artifacts due to voxel discretization.

\textbf{Implicit Surfaces:}
In contrast to explicit reconstruction methods that produce surface elements directly, several approaches have investigated the use of implicit representations. The most well-known example involves the Signed Distance Function (SDF) method, which discretizes space into voxels that encode the distance to the nearest surface. Pioneered by Hoppe et al. \cite{hoppe_surface_1992} and popularized by systems like KinectFusion \cite{newcombe_kinectfusion_2011}, SDF-based methods build implicit reconstructions that support real-time performance. More recent approaches, such as VoxBlox \cite{oleynikova_voxblox_2017} and VDBFusion \cite{vizzo_vdbfusion_2022}, extend these ideas to larger environments. Like occupancy-based methods, SDF representations require a post-processing step for surface extraction, making them less suitable for real-time applications requiring immediate surface access.

\begin{figure*}[ht!]
    \centering
    \includegraphics[width=1.0\textwidth]{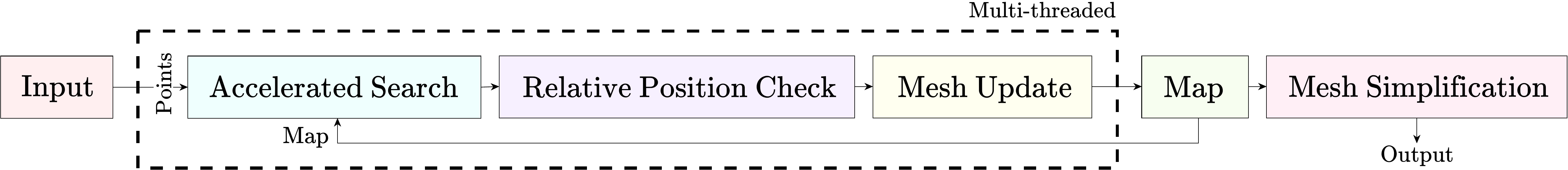}
    \caption{System architecture overview. The system processes input points in parallel to reconstruct a map, which is then simplified before output. Key components include Accelerated Search (Sec.~\ref{sec:accelerated_search}), Relative Position Check (Sec.~\ref{sec:relative_position_check}), Mesh Update (Sec.~\ref{sec:mesh_update}), and Mesh Simplification (Sec.~\ref{sec:simplify_mesh}).}
    \label{fig:system_overview}
    \vspace{-0.3cm}
\end{figure*}

\textbf{Global Interpolation:}
Instead of using input points to generate new geometry to represent a scene, global interpolation methods make direct estimates of the missing surface between the points. Poisson reconstruction \cite{kazhdan_poisson_2006} is a widely used method of this type, solving dense linear systems to recover smooth, high-quality surfaces from point clouds. Recent efforts, such as PUMA \cite{vizzo_poisson_2021}, modify Poisson reconstruction to improve efficiency, making it more suitable for real-time processing while still leveraging its surface estimation capabilities.
However, due to its global implicit nature, Poisson reconstruction tends to hallucinate geometry in sparse regions, particularly along open sky boundaries. While fine-tuning parameters like reconstruction depth and output density can help, the method is highly sensitive to these choices and often requires different settings depending on the scale of the object being reconstructed.

\section{Method}
\label{sec:Method}

\subsection{System Overview}

The input to PlanarMesh is a stream of 3D point cloud scans from a mobile LiDAR sensor. We assume that an accurate pose for each scan has been estimated by a separate LiDAR odometry system (such as FastLIO~\cite{xu_fast-lio2_2022} or VILENS~\cite{wisth_vilens_2023}) which includes motion correction for its continuous scanning pattern. 
Each point from the point cloud is then processed by our system in a multi-threaded manner. Each point is queried against two separate search trees built from the point data—Face Intersection Search (FIS) and Reverse Radius Search (RRS)— to identify candidate planar-meshes for updates. A relative position check determines the appropriate mesh update operation. Before outputting a mesh map file, a mesh simplification step reduces file size by eliminating redundant triangulations.
Fig.~\ref{fig:system_overview} provides a graphical overview of the system, and the algorithm's pseudocode is presented in Algo.~\ref{algo:main} in Sec.~\ref{sec:mesh_update}.

\subsection{Planar-Mesh Representation}
\label{sec:representation}

In our system, the surface of an object or building is modeled as a collection of \textbf{planar-meshes} $\map:\{\planarmesh^0, \planarmesh^1, ..., \planarmesh^N\}$ (see Fig.~\ref{fig:relative_position_check}). Each planar-mesh $\planarmesh$ consists of two components:
\begin{enumerate} 
    \item \textbf{Plane} $\plane$, represents a surface patch, parameterized by its position $\planePosition$ and a normal vector $\planeNormal$, computed from LiDAR samples $\planePoints \in \plane$.
    
    \item \textbf{Mesh} $\mesh$, %
     a triangular mesh composed of vertices $\vertices$, edges $\edges$ and faces $\faces$, all of which lie on the associated plane $\plane$. Each $\vertex \in \vertices$ has an associated radius $\radius$ that captures the local curvature.

\end{enumerate}

We chose a planar representation for reconstruction because the built environment largely consists of planar patches, such as walls, ceilings, and floors.
Most reconstruction approaches make prior assumptions~\cite{berger_survey_2017}. A prior that favors smooth low-frequency surfaces cannot accurately model complex geometries, while a prior which seeks to model high-frequency surfaces can result in  overfitting to the noise in the sensor measurements.
To address this trade-off, our system is based on planes, which provide a simple yet robust surface approximation that minimizes sensitivity to noise, and uses a collection of planar-meshes (planes with boundaries) to represent complex geometries that can't be represented by a single plane. Although not implemented in this work, we believe such a representation is well-suited for downstream applications such as floor plan generation and 3D building information modeling (BIM). Moreover, this representation maintains a level of granularity comparable to point- or surfel-based methods, which allows for potential surface adjustments via elastic deformation in future work.

\subsection{Accelerated Search}
\label{sec:accelerated_search}

\begin{figure}[h!]
    \centering
    \vspace{-0.3cm}
    \includegraphics[width=1.0\linewidth]{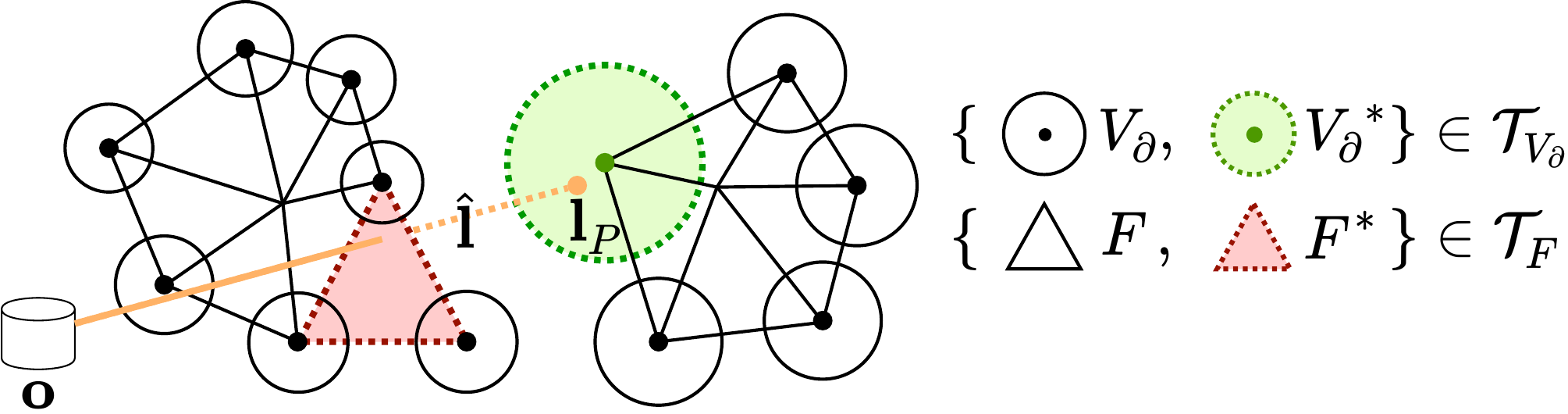}
    \caption{The two BVHs used in our system. Given an input ($\lidarRay$, $\lidarPoint$), the two BVHs retrieve intersected faces $\intersectingFaces$ and boundary vertices $\includingVertices$ that contain $\lidarPoint$ within their radius (a smaller sphere is used for display purposes).}
    \label{fig:connectivity_testing}
    \vspace{-0.3cm}
\end{figure}

Given the input LiDAR point $\lidarPoint$, sensor origin $\lidarOrigin$ and the corresponding ray $\lidarRay = (\lidarPoint - \lidarOrigin) / \|\lidarPoint - \lidarOrigin\|)$, it is essential to make efficient queries of our existing planar-mesh model $\planarmesh$ to integrate this new information into our reconstruction. To achieve this, we construct two Bounded Volume Hierarchies (BVH) trees (see Fig.~\ref{fig:connectivity_testing}):

\begin{itemize}
    \item Face Intersection Search Tree $\FISTree$: Stores all the triangular faces $\faces \in \mesh$ and returns a set of intersected faces $\intersectingFaces$ given the ray $\lidarRay$. This result is used to integrate free-space-information into our reconstruction (see Sec.~\ref{sec:mesh_update/integration_of_free_space_information}).
    \item Reverse Radius Search Tree $\RRSTree$: Stores all boundary vertices $\boundaryVertices \in \mesh$. Given a query point $\lidarPoint$, it returns the set of vertices $\includingVertices$ that contain $\lidarPoint$ within its associated radius $\radiusBoundary$. Since the radii $\radiusBoundary$ vary dynamically across $\boundaryVertices$, performing a reverse search using $\RRSTree$—which accounts for these varying radii—is more efficient than using a traditional KD-Tree with a large fixed radius followed by post-filtering. This result is used to guide our adaptive-resolution meshing (see Sec.~\ref{sec:mesh_update/shrink}).
\end{itemize}

Given the incremental update process of our system, the two BVH trees must be frequently updated to reflect changes in our mesh reconstruction. However, existing off-the-shelf BVH libraries, such as Embree~\cite{wald_embree_2014} and OptiX~\cite{parker_optix_2010}, do not support incremental insertions and deletions, instead requiring a full search tree to be rebuild with each update. While individual rebuilds are computationally efficient, performing several thousand rebuilds per scan would introduce a substantial overhead, making this approach impractical. Instead, we developed a custom accelerated BVH structure in accordance to Erin Catto’s 2019 GDC talk\footnote{Erin Catto, GDC 2019 talk ``Math for Game Developers: Dynamic Bounding Volume Hierarchies", \url{https://gdcvault.com/play/1025909/Math-for-Game-Developers-Dynamic}.}. Our implementation adopts the dynamic insertion strategy described in the talk, which helps avoid costly full tree rebuilds. Additionally, we implemented multi-threaded query support using fine-grained locking to fully parallelize the mesh-construction pipeline.

\subsection{Relative Position Check}
\label{sec:relative_position_check}

After querying the search trees, $\FISTree$ and $\RRSTree$, with $\lidarRay$ and $\lidarPoint$, we obtain lists of $\intersectingFaces$ and $\includingVertices$ whose associated planar-meshes $\planarmeshes$ will be updated or modified using the $\lidarPoint$ (see Sec.~\ref{sec:mesh_update}). Prior to updating the planar-meshes $\planarmeshes$, the relative position between $\lidarPoint$ and each $\planarmesh \in \planarmeshes$ must be known to select the correct operation to be performed on $\planarmesh$.

\begin{figure}[h!]
    \centering
    \includegraphics[width=1.0\linewidth]{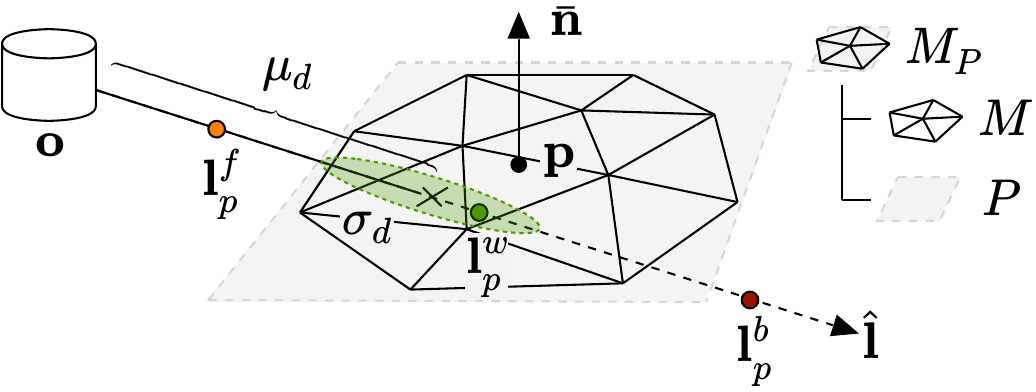}
    \caption{Illustration of planar-mesh $\planarmesh$. The relative position of $\lidarPoint$ with respect to $\planarmesh$ is determined by comparing its location against the uncertainty threshold $\sigma_d$, visualized as an ellipse.}
    \label{fig:relative_position_check}
\end{figure}

A lidar point $\lidarPoint$ can be related to a planar-mesh $\planarmesh$ in three different ways (see Fig.~\ref{fig:relative_position_check}):

\begin{enumerate}
    \item $\lidarPoint$ is observed in front of the $\planarmesh$, denoted as $\lidarPoint^{\text{f}}$ and $\planarmeshFront$.
    \item $\lidarPoint$ is observed within a threshold $\sigma_d$ from the $\planarmesh$, denoted as $\lidarPoint^{\text{w}}$ and $\planarmeshWithin$ (typically because it lies on an underlying surface the plane $\plane \in \planarmesh$ approximates).
    \item $\lidarPoint$ is observed behind the $\planarmesh$, denoted as $\lidarPoint^{\text{b}}$ and $\planarmeshBehind$.
\end{enumerate}

The position of $\lidarPoint$ relative to $\planarmesh$ can be found by comparing the measured range $\range = \|\lidarPoint - \lidarOrigin\|_{2}$ to the expected range $\expectedDistance = (\planePosition - \lidarOrigin) \cdot \planeNormal / (\planeNormal \cdot \lidarRay)$ using z-score $z = (\range - \expectedDistance)/\stdDistance$, where $\stdDistance^2 = \sigma_{\planePosition}^2 + \sigma_{\lidarOrigin}^2 + \sigma_{\planeNormal}^2 + \sigma_{\lidarRay}^2 + \sigma_{\text{noise}}^2$ with sensor noise $\sigma_{\text{noise}}$ along the range axis. We compute each $\sigma_{\{\cdot\}}^2 = \mathbf{J}_{\{\cdot\}} \, \mathbf{\Sigma}_{\{\cdot\}} \, \mathbf{J}_{\{\cdot\}}^{\top}$, where $\mathbf{J}_{\{\cdot\}}$:
\begin{align*}
    \mathbf{J}_{\planePosition} &= \frac{\partial \expectedDistance}{\partial \planePosition} 
    = \frac{\planeNormal}{\planeNormal \cdot \lidarRay}, 
    &\mathbf{J}_{\planeNormal} &= \frac{\partial \expectedDistance}{\partial \planeNormal} 
    = \frac{\lidarPoint - \lidarOrigin}{\planeNormal \cdot \lidarRay} 
    - \expectedDistance \frac{\lidarRay}{\planeNormal \cdot \lidarRay}, 
    \\
    \mathbf{J}_{\lidarOrigin} &= \frac{\partial \expectedDistance}{\partial \lidarOrigin} 
    = -\frac{\planeNormal}{\planeNormal \cdot \lidarRay}, 
    &\mathbf{J}_{\lidarRay} &= \frac{\partial \expectedDistance}{\partial \lidarRay} 
    = -\expectedDistance \frac{\planeNormal}{\planeNormal \cdot \lidarRay}.
\end{align*}

We then perform a hypothesis test using a 95\% confidence interval, corresponding to a critical value of 1.96:
\begin{equation*}
    \begin{aligned}
        H_0 &: \phantom{|}z\phantom{|} < -1.96,  &\quad &\lidarPoint^{\text{f}} \text{ observed in front of $\planarmesh$}, \\
        H_1 &: |z| \leq \phantom{-}1.96,  &\quad &\lidarPoint^{\text{w}} \text{ observed within $\planarmesh$}, \\
        H_2 &: \phantom{|}z\phantom{|} > \phantom{-}1.96,  &\quad &\lidarPoint^{\text{b}} \text{ observed behind $\planarmesh$},
    \end{aligned}
\end{equation*}

to classify the relation into the three cases mentioned above.

\subsection{Mesh Update}
\label{sec:mesh_update}

\begin{figure}[h!]
    \vspace{-0.6cm}
    \centering
    \includegraphics[width=1.0\linewidth]{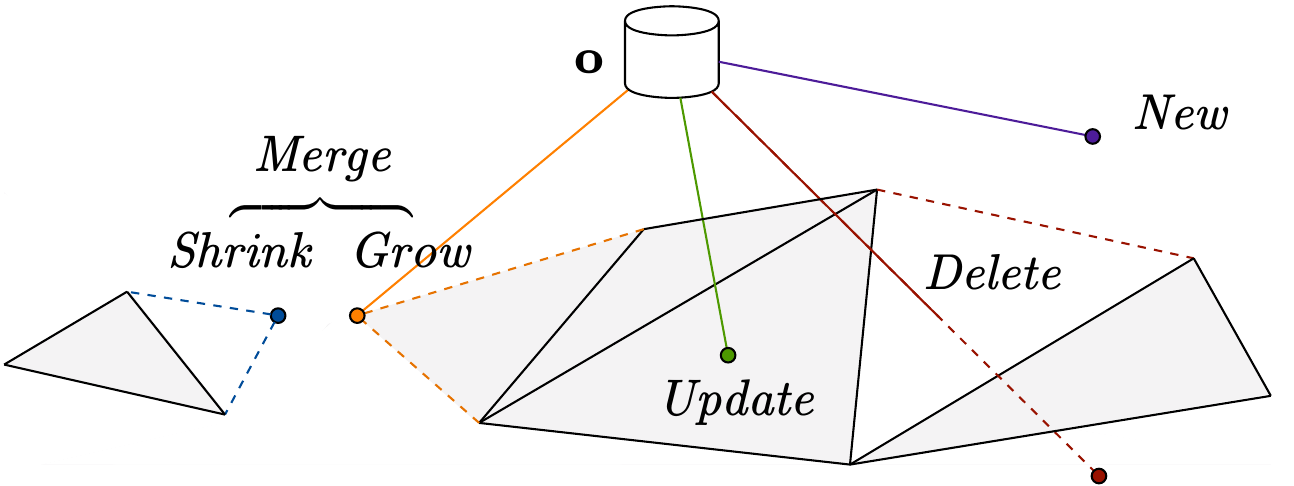}
    \caption{Different mesh update operations: \textit{Update (Sec.~\ref{sec:mesh_update/update})}, \textit{Grow (Sec.~\ref{sec:mesh_update/grow})}, \textit{New (Sec.~\ref{sec:mesh_update/new})}, \textit{Delete (Sec.~\ref{sec:mesh_update/delete})}, \textit{Shrink (Sec.~\ref{sec:mesh_update/shrink})}, \textit{Merge (Sec.~\ref{sec:mesh_update/merge}}).}
    \label{fig:mesh_update_operations}
\end{figure}

Let $\planarMeshesOfFaces$, $\planarMeshesOfVertices$ denote the planar-meshes associated with $\intersectingFaces$, $\includingVertices$ returned from querying $\FISTree$, $\RRSTree$ respectively. After the relative position check (Sec.~\ref{sec:relative_position_check}), we classify the planar-meshes into three groups for each $\intersectingFaces$, $\includingVertices$: $\planarmeshes_{\{\cdot\}}^{\text{front}}$, $\planarmeshes_{\{\cdot\}}^{\text{within}}$, $\planarmeshes_{\{\cdot\}}^{\text{behind}}$. Additionally, we define $\planarmeshes_{\{\cdot\}}^{\text{seed}}$ to contain planar-mesh that does not have sufficient points for relative position check to be performed (more in Sec.~\ref{sec:update_mesh/seed_planar-mesh_initialization}). The updates and modifications performed on $\planarmesh$ in these groups are illustrated in Fig.~\ref{fig:mesh_update_operations}, with the corresponding algorithm provided in Algo.~\ref{algo:main} and described below:

\begin{algorithm}
    \caption{Incremental PlanarMesh Reconstruction}
    \label{algo:main}
    \begin{algorithmic}[1]
        \Require Map $\map$, LiDAR Point $\lidarPoint$, sensor origin $\lidarOrigin$, 
        \Ensure Updated Map $\updatedMap$

        \Comment{Accelerated Search}
        \State $\intersectingFaces, \planarMeshesOfFaces \gets \FISFunc(\FISTree,\lidarRay)$
        \State $\includingVertices, \planarMeshesOfVertices \gets \RRSFunc(\RRSTree, \lidarPoint)$

        \Comment{Relative Position Check}
        \State $\planarMeshesOfFacesWithin, \planarMeshesOfFacesBehind \gets \{\planarmesh \mid \texttt{RPC}(\planarmesh,\lidarPoint,\lidarOrigin),\, \planarmesh \in \planarMeshesOfFaces\}$
        \State $\planarMeshesOfVerticesWithin, \planarMeshesOfVerticesSeed \gets \{\planarmesh \mid \texttt{RPC}(\planarmesh,\lidarPoint,\lidarOrigin),\, \planarmesh \in \planarMeshesOfVertices\}$
        
        \Comment{Mesh Update}
        
        \Comment{\textit{Update (Sec.~\ref{sec:mesh_update/update})}}
        \If{$\planarMeshesOfFacesWithin \neq \emptyset$} \label{line:start_of_update}
            \State $\planarmesh^{*} \gets \arg\max_{\planarmesh \in \planarMeshesOfFacesWithin} \texttt{Area}(\planarmesh)$
            \State $\updatedPlanarMesh \gets \texttt{UPDATE}(\planarmesh^{*}, \lidarPoint)$
        \label{line:end_of_update}

        \Comment{\textit{Grow (Sec.~\ref{sec:mesh_update/grow})}}
        \ElsIf{$\planarMeshesOfVerticesWithin \neq \emptyset$}
        \label{line:start_of_grow}
            \State $\planarmesh^{*} \gets \arg\max_{\planarmesh \in \planarMeshesOfVerticesWithin} \texttt{Area}(\planarmesh)$
            \State $\updatedPlanarMesh \gets \texttt{GROW}(\planarmesh^{*}, \lidarPoint)$
        \ElsIf{$\planarMeshesOfVerticesSeed \neq \emptyset$}
            \State $\planarmesh^{*} \gets \arg\min_{\planarmesh \in \planarMeshesOfVerticesSeed} \texttt{Dist}(\planarmesh, \lidarPoint)$
            \State $\updatedPlanarMesh \gets \texttt{GROW}(\planarmesh^{*}, \lidarPoint)$
        \label{line:end_of_grow}

        \Comment{\textit{New (Sec.~\ref{sec:mesh_update/new})}}
        \Else 
        \label{line:start_of_new}
            \State $\newSeedPlanarMesh \gets \texttt{NewSeedPlanarMesh}(\lidarPoint)$
        \EndIf \label{line:end_of_new}

            \Comment{\textit{Delete (Sec.~\ref{sec:mesh_update/delete})}}
            
        \If{$\planarmesh^{*} \in \{\planarMeshesOfFacesWithin,\, \planarMeshesOfVerticesWithin\}$} \label{line:start_of_additional}
        \label{line:start_of_delete}

            \State $\updatedPlanarMeshesOfFacesBehind \gets \texttt{DELETE}(\intersectingFaces \in \planarMeshesOfFacesBehind)$
            \label{line:end_of_delete}

            \Comment{\textit{Shrink (Sec.~\ref{sec:mesh_update/shrink})}}
            \State $\updatedPlanarMeshesOfVertices \gets 
            \texttt{SHRINK}( \includingVertices \in \{\planarMeshesOfVertices \setminus \planarmesh^{*}\}, \lidarPoint)$
            \label{line:end_of_shrink}
        \EndIf

        \\
        \State $\updatedMap \gets \map \cup \updatedPlanarMesh \cup \newSeedPlanarMesh \cup \updatedPlanarMeshesOfFacesBehind \cup \updatedPlanarMeshesOfVertices$
        
    \end{algorithmic}
\end{algorithm}

\subsubsection{Update (Line~\ref{line:start_of_update} to \ref{line:end_of_update})}
\label{sec:mesh_update/incremental_plane_fitting}
\label{sec:mesh_update/update}

If $\planarMeshesOfFacesWithin \neq \emptyset$, then we update the plane parameters $(\planePosition, \planeNormal)$ of the largest planar-mesh $\planarmesh^{*} \in \planarMeshesOfFacesWithin$. This is achieved by performing Principle Component Analysis (PCA) on the covariance matrix $\Sigma_{\plane}$ of $\planePoints \in \plane$ and $\lidarPoint$, and choosing the eigenvector corresponding to the smallest eigenvalue as the updated plane normal $\planeNormal$.

Instead of re-computing $\Sigma_{\plane}$ from scratch each time given $\planePoints$ and $\lidarPoint$, we incrementally update $\Sigma_{\plane} \gets \tilde{\Sigma}_{\plane}$:
\begin{align}
    \tilde{\Sigma}_{\plane} = \frac{n_{\plane} \Sigma_{\planePosition} + \Sigma_{\lidarPoint} + n_{\plane}\Delta_{\planePosition}\Delta_{\planePosition}^\top + \Delta_{\lidarPoint}\Delta_{\lidarPoint}^\top}{1 + n_{\plane}},
\end{align}
where $n_{\plane}$ is size of $\planePoints$, $\Delta_{\{\cdot\}} = \{\cdot\} - \mu$ and $\mu = (\lidarPoint + n_{\plane} \cdot \planePosition)/(1+n_{\plane})$. We discuss the reason behind choosing the largest planar-mesh in Sec.~\ref{sec:merging_effect}.

\subsubsection{Grow (Line~\ref{line:start_of_grow} to \ref{line:end_of_grow})}
\label{sec:mesh_update/mesh_expansion}
\label{sec:mesh_update/grow}

If $\planarMeshesOfVerticesWithin \neq \emptyset$, then we expand the largest planar-mesh $\planarmesh^{*} \in \planarMeshesOfVerticesWithin$, otherwise we expand the closest seed planar-mesh $\planarmesh^{*} \in \planarMeshesOfVerticesSeed$.
The \textit{Grow} step has lower priority over the \textit{Update} step as an expansion creates mesh elements in newly observed space, which is less certain than updating an existing mesh we had previously observed. We expand $\planarmesh^{*}$ by iteratively creating edges $\tilde{\edges}$ between $\lidarPoint$ and $\includingVertices \in \planarmesh^{*}$ provided that $\tilde{\edge} \in \tilde{\edges}$ does not intersect with $\edges \in \planarmesh^{*}$. Additionally, we create new faces $\tilde{\faces}$ between two consecutive edges in $\tilde{\edges}$ if $\tilde{\face} \in \tilde{\faces}$ does not contain an existing $\vertex \in \planarmesh^{*}$.

\subsubsection{New (Line~\ref{line:start_of_new} to \ref{line:end_of_new})}
\label{sec:update_mesh/seed_planar-mesh_initialization}
\label{sec:mesh_update/new}
We create new seed planar-meshes $\newSeedPlanarMesh$ if $\lidarPoint$ is added to an empty region of the map. $\newSeedPlanarMesh$ is later updated or expanded as described in previous two steps. A planar-mesh is consider seed only if it has less than three points, or has a small mesh area, as it results in unstable plane fitting due to noisy LiDAR point $\lidarPoint \in \planePoints$.

\subsubsection{Delete (Line~\ref{line:start_of_delete} to \ref{line:end_of_delete})}
\label{sec:mesh_update/integration_of_free_space_information}
\label{sec:mesh_update/delete}

If the new point $\lidarPoint$ is added to a planar-mesh $\planarmesh^{*}$ considered ``within", i.e. $\planarmesh^{*} \in \{\planarMeshesOfFacesWithin,\, \planarMeshesOfVerticesWithin\}$, $\lidarPoint$ is likely to be an inlier point and the corresponding ray $\lidarRay$ can used to integrate free-space information. 
We remove all faces $\faces$ whose planar-meshes have been ``passed through'' by $\lidarPoint$ i.e. $\faces \in \planarMeshesOfFacesBehind$. An edge $\edge \in \planarMeshesOfFacesBehind$ with no connected faces is considered ``dangling'' and is deleted from $\planarMeshesOfFacesBehind$. This mechanism eliminates any faces and edges that erroneously span gaps between adjacent surface in the reconstructed map.

The fidelity of these ``pass through" checks improves as more points are added to a plane, which in turn results in better plane estimates with lower $\Sigma_{\planePosition}$ and $\Sigma_{\planeNormal}$. This, in turn, reduces the range uncertainty $\stdDistance$ used in the relative position check between a point $\lidarPoint$ and a $\planarmesh$, enabling more effective identification of “pass-through” points. In practice this allows subtle features to be captured more reliably, such as the slightly recessed window panes in a panel door (see Fig.~\ref{fig:compare_results} magenta box) or chamfered edges in walls.

\subsubsection{Shrink (Line~\ref{line:end_of_shrink})}
\label{sec:mesh_update/adaptive_resolution_meshing}
\label{sec:mesh_update/shrink}

For $\planarmesh^{*} \in \{\planarMeshesOfFacesWithin,\, \planarMeshesOfVerticesWithin\}$, we integrate $\lidarPoint$ into the map to shrink nearby planar-meshes with high curvature and long edges, encouraging their replacement with smaller, more localized triangles.

To achieve this, we introduce the concept of the Reverse Radius Search (RRS), where the radius value approximates the medial axis distance—defined as the shortest distance from a boundary point to its medial axis—which is known to correlate with local surface curvature~\cite{amenta_new_1998}. Specifically, we define the radius $\radius$ as the shortest distance from a vertex $\vertex \in \planarmesh^{\text{current}}$ to its nearest neighbor $\vertex_{\text{nearest}} \in \planarmesh^{\text{nearest}}$ from a different planar-mesh.

When $\lidarPoint$ is added to $\planarmesh^{*}$ as a new vertex, it updates the radius of nearby boundary vertices $\includingVertices \in \planarmeshes^{\text{other}}$, where $\planarmeshes^{\text{other}} \gets \{\planarMeshesOfVertices \setminus \planarmesh^{*}\}$. Any edges connected to $\includingVertices$ that exceed its updated radius are removed. A vertex $\vertex \in \includingVertices$ with no remaining edges is considered ``dangling'' and is deleted from $\planarmeshes^{\text{other}}$, though it may later be reintroduced as a new LiDAR point.
This process allows other LiDAR points that are closer to $\planarmeshes^{\text{other}}$ to be incorporated with edges that conform to the local radius threshold. As a result, high-curvature regions (typically near object edges) are refined with smaller triangles, improving geometric accuracy. Conversely, in low-curvature regions, larger triangles facilitate rapid surface expansion with fewer sampled points, yielding a more compact yet efficient representation.

\subsubsection{Merge}
\label{sec:merging_effect}
\label{sec:mesh_update/merge}

The final operation seeks to merge two planar-meshes $\planarmesh^{\text{large}}, \planarmesh^{\text{small}}$ which are close to each other and have similar orientations. When a new point $\lidarPoint$ is integrated, we prioritize adding it to the larger mesh $\planarmesh^{\text{large}}$. As $\planarmesh^{\text{large}}$ expands, the radius of $\boundaryVertices \in \planarmesh^{\text{small}}$ is reduced, triggering a removal of edges $\edges \in \planarmesh^{\text{small}}$ in the smaller mesh (\textit{Shrink}), and subsequent addition of the ``dangling" vertices to the larger mesh $\planarmesh^{\text{large}}$ (\textit{Grow}). This results in $\planarmesh^{\text{large}}$ ``consuming'' $\planarmesh^{\text{small}}$, creating an spontaneous merge effect.

\subsection{Mesh Simplification}
\label{sec:simplify_mesh}

Before finalizing the map $\map$ and output it as a file, we simplify $\planarmesh \in \map$ as follows:
\begin{enumerate}
    \item Collect all vertices $\vertices$ within the planar-mesh $\planarmesh$.
    \item Sort $\vertex \in \vertices$ in ascending order based on their radius $\radius$.
    \item Traverse the sorted list, add a vertex $v$ to $\vertices_{\text{sampled}}$ only if no previously sampled vertices $\vertex_{\text{sampled}} \in \vertices_{\text{sampled}}$ exist within its radius $\radius$.
    \item Perform Delaunay triangulation on the sampled vertices $\vertices_{\text{sampled}}$ to generate a convex mesh $\mesh_{\text{convex}}$.
    \item Restore concavity by discarding $\faces_{\text{convex}} \in \mesh_{\text{convex}}$ whose centroids do not fall within an original face $\faces \in \planarmesh$.
\end{enumerate}

\section{Experimental Results}
\label{sec:experiments}

\subsection{Experimental Setup and Evaluation}
\label{ssec:exp_setup}

\begin{figure*}[h!]
	\centering
	\includegraphics[width=\linewidth]{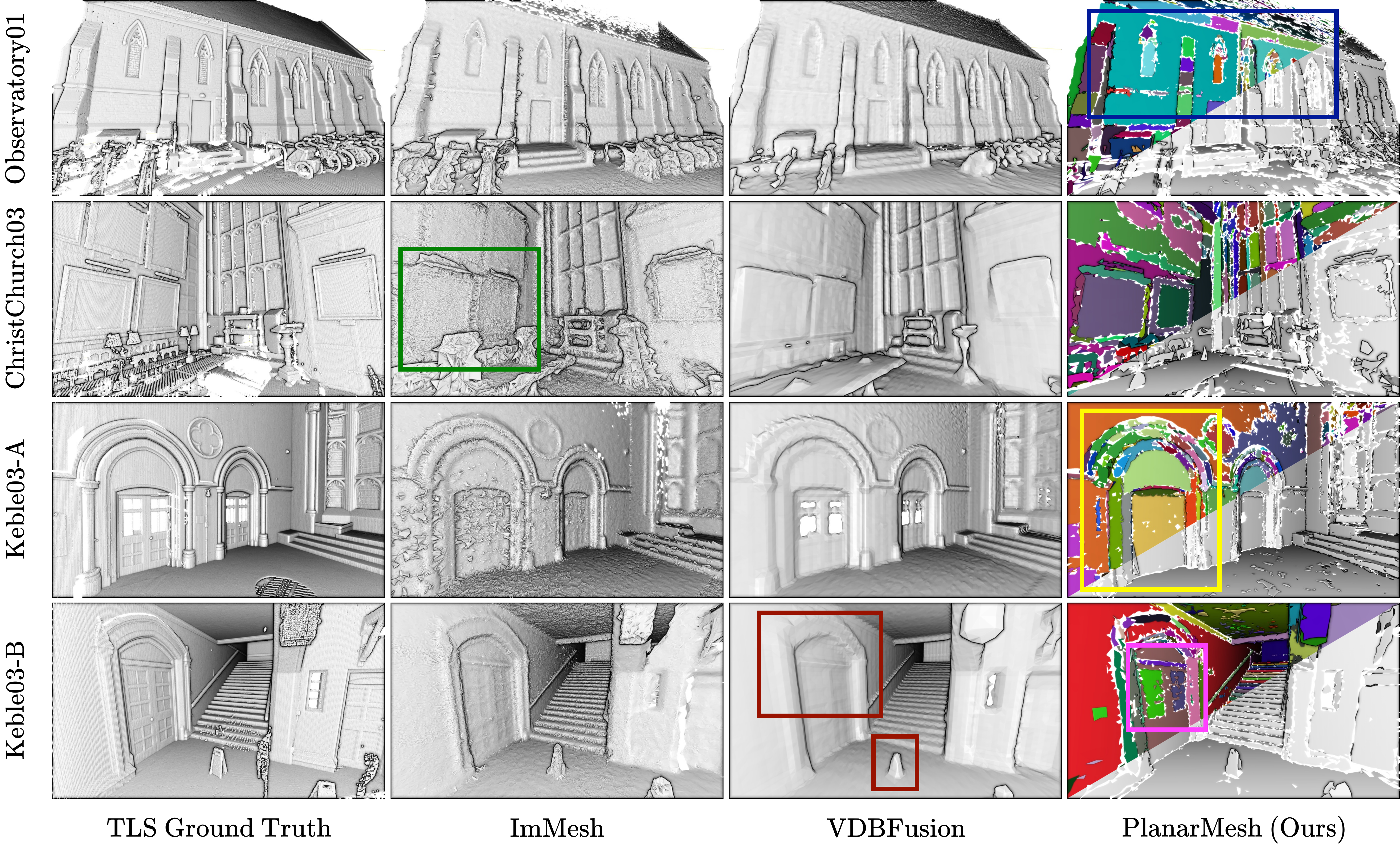}
	\caption{Comparison between reconstructions from the different methods. Individual planar-meshes produced by PlanarMesh are colored. It can correctly detect subtle features such as the geometry around a door frame (yellow), recessed window panes in a paneled door (magenta) while fitting a single large plane to a uniform wall (blue). Examples of noisy surfaces from ImMesh (green) and overly-smoothed surfaces from VDBFusion (red) are also indicated.}
    \label{fig:compare_results}
    \vspace{-0.3cm}
\end{figure*}

We evaluate our reconstruction method using the Oxford Spires dataset~\cite{tao2024spires} across diverse indoor and outdoor environments. The sequences used are listed in Tab.~\ref{tab:reconstruction_eval}. We compare our method to the following approaches: ImMesh~\cite{lin_immesh_2023} (voxel mesh-based), VDBFusion~\cite{vizzo_vdbfusion_2022} (TSDF-based), and OctoMap~\cite{hornung_octomap_2013} (occupancy-based). Although OctoMap is not a mesh-based reconstruction method, we include it as a reference point: it provides a theoretical upper bound on recall and offers insight into the compression capabilities of mesh-based representations. All methods are configured for real-time mesh generation at approximately 1 Hz, achieved by setting the voxel size to 0.05~m for OctoMap and 0.1~m for both ImMesh and VDBFusion, running on a single CPU core. At normal walking speeds this corresponds to adding new scan every 0.5\,-1\,m traveled.

For each approach, we process individual LiDAR scans with ground truth poses to produce mesh reconstructions. The ground truth poses were obtained by registering each individual undistorted point cloud scans to the highly accurate TLS point cloud map of the test location.

We performed quantitative evaluation by converting reconstructed meshes into point clouds, uniformly sampling points to match the raw LiDAR scan count. To evaluate reconstruction quality, we report the mean and standard deviation of the distances to the ground truth, alongside precision, recall, and F-score at a 0.1\,m threshold (Tab.~\ref{tab:reconstruction_eval}). We applied pre-filtering, as detailed in~\cite{tao2025silvr}, to both the ground truth TLS scan and the reconstructed meshes, ensuring a fair comparison by excluding areas not captured by both scanning modalities. We also report the file size
of the output mesh (in PLY binary format). All experiments were conducted on a 28-core Intel i7 CPU without GPU acceleration. While all baseline methods were run on a single core, our method requires all available CPU cores to achieve the aforedmentioned real-time performance.

\begin{table*}[h!]
	\caption{\small{Evaluation of 3D Reconstruction Quality. OctoMap is placed as its own category as it is an occupancy-based method and therefore does not produce face or vertex counts.}}
   	\setlength{\tabcolsep}{3pt} %
	\centering
	\begin{tabular}{ l c r r r c c c c c}
    
		\toprule
        
		Method & Per-Scan Time$\downarrow$ & File size$\downarrow$ & Num of Faces$\downarrow$ & Num of Vertices$\downarrow$ & Mean$\downarrow$ & Std$\downarrow$ &  Precision$\uparrow$ & Recall$\uparrow$  & F-Score$\uparrow$ \\
        & (s) & (MB) &  &  & (m) & (m) &  &  & \\
         
        \hline \addlinespace
        
		\textbf{Christ Church 03} ($\sim$307 m)\\
        
		\hline \addlinespace

            VDBFusion & 0.871 & 53.6 & 1,992,391 & 1,152,788 & 0.044 & \textbf{0.077} & 0.918 & 0.970 & 0.943 \\
            ImMesh & 0.724 & 370.9 & 21,180,823 & 7,959,789 & 0.090 & 0.186 & 0.820 & 0.990 & 0.897 \\
            PlanarMesh (Ours) & \textbf{0.392} & 10.1 & \textbf{398,712} & \textbf{411,907} & \textbf{0.037} & 0.081 & \textbf{0.951} & 0.964 & 0.957 \\
            \cdashline{1-10}
            OctoMap & 0.432 & \textbf{3.4} & N/A & N/A & 0.040 & 0.083 & 0.943 & \textbf{0.991} & \textbf{0.966} \\

        \hline \addlinespace
        
		\textbf{Keble College 03} ($\sim$108 m)\\
        
		\hline \addlinespace

            VDBFusion & 0.968 & 51.3 & 1,821,087 & 1,150,250 & 0.033 & 0.113 & 0.962 & 0.940 & 0.951 \\
            ImMesh & 0.355 & 163.8 & 9,057,110 & 3,838,040 & 0.035 & \textbf{0.064} & 0.955 & 0.918 & 0.936 \\
            PlanarMesh (Ours) & 0.416 & \textbf{7.3} & \textbf{287,020} & \textbf{296,254} & \textbf{0.031} & 0.134 & \textbf{0.979} & 0.894 & 0.935 \\
            \cdashline{1-10}
            OctoMap & \textbf{0.328} & 24.3 & N/A & N/A & 0.040 & 0.159 & 0.964 & \textbf{0.966} & \textbf{0.965} \\
                
        \hline \addlinespace
        
		\textbf{Observatory 01} ($\sim$324 m)\\
        
		\hline \addlinespace

            VDBFusion & 2.406 & 148.5 & 5,246,193 & 3,346,024 & 0.047 & 0.104 & 0.899 & 0.899 & 0.899 \\
            ImMesh & 0.448 & 424.7 & 23,448,665 & 9,986,250 & 0.056 & \textbf{0.089} & 0.878 & 0.832 & 0.854 \\
            PlanarMesh (Ours) & \textbf{0.213} & \textbf{15.3} & \textbf{546,415} & \textbf{682,725} & \textbf{0.042} & 0.114 & \textbf{0.929} & 0.847 & 0.886 \\
            \cdashline{1-10}
            OctoMap & 0.659 & 67.8 & N/A & N/A & 0.055 & 0.150 & 0.896 & \textbf{0.941} & \textbf{0.918} \\
                
		\bottomrule
        \vspace{-0.6cm}

	\end{tabular}
	\footnotesize{} 
	\label{tab:reconstruction_eval}
\end{table*}

\subsection{Mesh Reconstruction Quality}
\label{ssec:mesh_reco_quality}

In this section, we evaluate the reconstruction quality of our approach. Fig.~\ref{fig:compare_results} provides a visual comparison between the mesh reconstructions from the different methods.
Both VDBFusion and Immesh use a fixed voxel size (0.1~m) for their reconstruction, whereas our approach adapts to scene curvature. This allows us to model complex door frames with multiple planes (yellow box) and uniform walls with single large plane (blue box). 
We also observe that VDBFusion tends to oversmooth shapes,  which results in a loss of detail such as the door frame and the sign on the floor (red boxes), whereas our approach captures those details. On the other hand, ImMesh produces noisier meshes (green box) due to per-voxel plane fitting, resulting in rougher surfaces. While our mesh reconstruction retains these detailed features, it suffers from small holes and rugged edges in the mesh. This could be improved by scanning the scene more densely or by applying smoothing or plane intersection operations in processing, which we aim to incorporate in future work.

Tab.~\ref{tab:reconstruction_eval} presents quantitative evaluation of our approach.  Our reconstruction has a mean mesh-to-point error in the range of 3-4\,cm, outperforming the baselines. By grouping points that belong to the same plane, our planar-mesh representation effectively averages out point noise, leading to a better performance. We also obtain high precision (0.92\,-0.97), recall (0.84\,-0.96), and F1 scores (0.91\,-0.96) for all the sequences which is either better or comparable to other methods. Our approach is intentionally biased toward higher precision and lower recall, as we do not retain infinitesimally small faces, which are unlikely to represent meaningful structural elements in our target application of building modeling.

\subsection{File Footprint}
\label{ssec:exp_file_footprint}

We recognize that the built environment is dominated by planes, and our approach leverages this to achieve a significant file size reduction as seen in file and mesh size columns in \tabref{tab:reconstruction_eval}. Our method produces triangular meshes with 280\,-550\,K faces, an order of magnitude lower than other methods (1.8\,-23\,M). Compared to the accumulated LiDAR point cloud ($\sim$100 MB), our reconstruction yields a file size of approximately 10 MB, representing a roughly tenfold compression. 

Traditional mesh-based reconstructions often generate large files due to the storage of numerous small triangles, even in low-curvature regions. Whereas, our planar-mesh approach shares normals among faces within the same plane. By re-sampling vertices and re-triangulating on demand, we achieve significantly smaller file sizes.

\subsection{Timing Analysis}
\label{sec:exp_timing_analysis}

\begin{figure}[h!]
    \centering
    \includegraphics[width=1.0\linewidth]{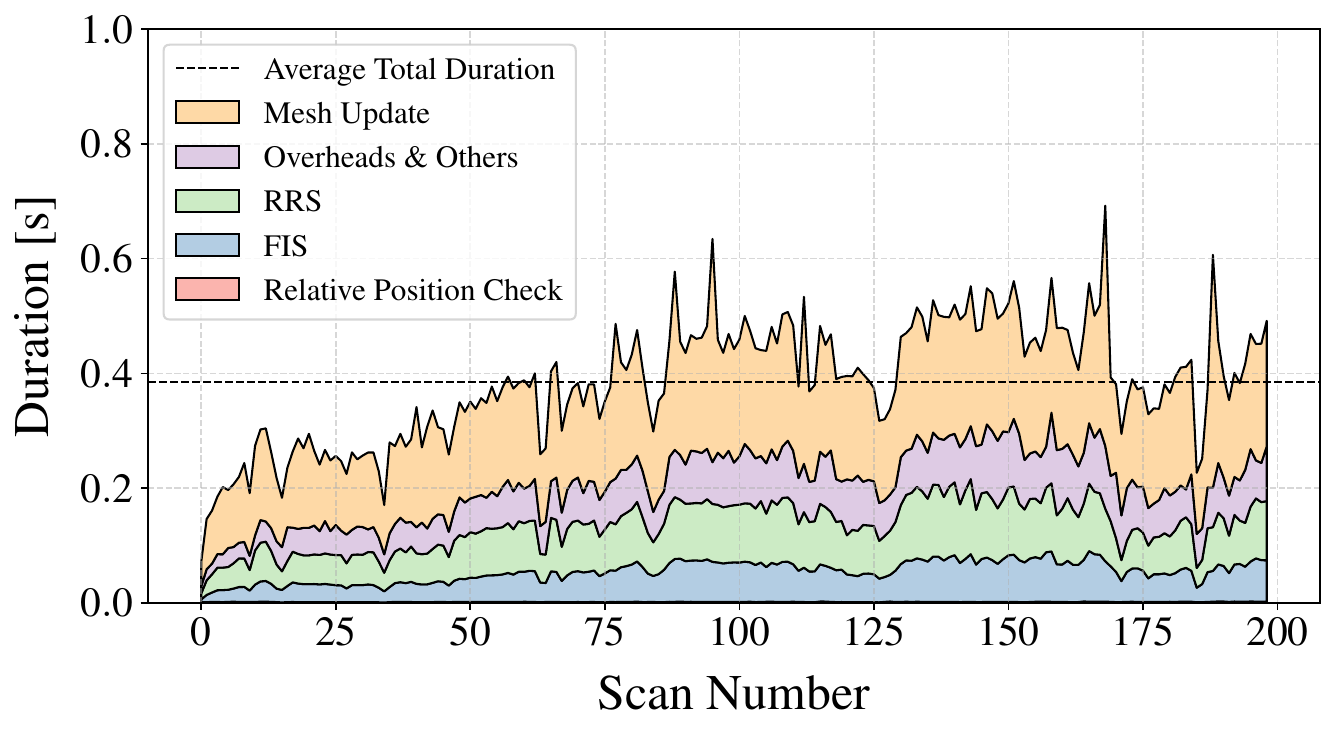}
    \caption{Computation time for each module of PlanarMesh, visualized as a stacked area plot for the ChristChurch03 sequence.}
    \label{fig:stack_area_duration_plot}
    \vspace{-0.3cm}
\end{figure}

This section validates the real-time capability of our approach and presents a breakdown of the computational time required by each module within our reconstruction pipeline. The processing time of each module is visualized as a stacked area plot (Fig.~\ref{fig:stack_area_duration_plot}). On average, each scan is processed in around 0.4\,sec, with peak processing times reaching 0.7\,sec, resulting in an effective processing speed of about \processingSpeed Hz. This performance is on-par with other real-time reconstruction methods (Tab.~\ref{tab:reconstruction_eval}). If desired, a higher update frequency could be achieved by down-sampling the input LiDAR point cloud appropriately - a viable strategy for applications demanding lower latency. The FIS and RRS data structures are implemented as binary search trees, the complexity of their search operations is therefore $\mathcal{O}(\log N)$, where $N$ is the number of faces or boundary vertices, respectively. 

\subsection{Ablation Studies}
\label{ssec:exp_ablation_size}

\begin{table}[h!]
	\caption{\small{Ablation Study on ChristChurch03 Dataset while varying the keep seed planar-meshes duration.}}
   	\setlength{\tabcolsep}{3pt} %
	\centering
	\begin{tabular}{ l c c c c c c c}
    
		\toprule
        
		Num & Time$\downarrow$ & File size$\downarrow$ & Mean$\downarrow$ & Std$\downarrow$ &  Precision$\uparrow$ & Recall$\uparrow$  & F-Score$\uparrow$ \\
        & (s) & (MB) & (m) & (m) &  &  & \\
         
        \hline \addlinespace

        0 & \textbf{0.295} & \textbf{5.9} & 0.037 & \textbf{0.075} & 0.946 & 0.889 & 0.917 \\
        1 & 0.347 & 7.9 & 0.036 & 0.076 & 0.951 & 0.946 & 0.948 \\
        10 & 0.384 & 8.3 & \textbf{0.036} & 0.076 & 0.951 & 0.951 & 0.951 \\
        All & 0.313 & 8.6 & 0.036 & 0.078 & \textbf{0.951} & \textbf{0.953} & \textbf{0.952} \\

		\bottomrule 
        \vspace{-0.6cm}

	\end{tabular}
	\footnotesize{} 
	\label{tab:ablation_studies}
\end{table}

We present the results of an ablation study on the retention duration of seed planar-meshes. Seed planar-meshes typically contain few points or have small surface areas, often arising from outlier points. However, they can also result from limited observations of complex geometries that have not yet been sufficiently scanned. In our implementation, we keep the seed planar-meshes throughout the duration of mission and do not discard them. By increasing the duration seed planar-meshes are kept, we enhance the likelihood of identifying and developing complex geometry, thereby improving the overall recall rate. However, this comes at the cost of longer processing times and larger file sizes, as demonstrated in Tab.~\ref{tab:ablation_studies}. By reducing the retention duration of the seed meshes from 10 scans to 1 scan, we are able to reduce the processing time from 0.384\, to 0.295\,sec and reduce the file size 8.6\, to 5.9\,MB. Notably, keeping all scans results in lower processing times by eliminating the overhead caused by the removal process, providing an extra benefit if file size is not prioritized. This trade-off can be achieved without any degradation in the precision.

\section{Conclusions}
In this work we introduced PlanarMesh, a novel real-time adaptive resolution mesh reconstruction system for 3D LiDAR data. By leveraging free-space information and introducing the Reverse Radius Search (RRS) for local curvature estimation, our method dynamically adjusts mesh resolution to capture both large-scale structures and fine detail efficiently. The system operates incrementally using multi-threaded BVH-based search operations, achieving real-time performance at approximately \processingSpeed Hz while significantly reducing memory footprint—producing mesh files up to \compressionRatio times smaller than raw input data.
Our experimental evaluation demonstrates that PlanarMesh achieves accuracy on par with, or exceeding state-of-the-art methods, with output file size more than 5~times smaller, making it ideal for scalable 3D mapping in robotics and mobile scanning applications. At present, the reconstruction system uses continually more memory until it exceeds available memory at about 300~scans (about 300\,m at 1\,m/sec walking). In future work we aim to introduce a submapping mechanism to offload older mesh components and to achieve bounded memory usage. Additionally, we aim to incorporate loop closure for improved consistency in large-scale mapping and further explore mesh hole filling and edge smoothing algorithms.

\bibliographystyle{IEEEtran}
\bibliography{references_handedited}

\end{document}